\documentclass[10pt,twocolumn,letterpaper]{article}

\usepackage{wacv}                         
\usepackage[accsupp]{axessibility} 
\usepackage{siunitx}
\usepackage{multirow}
\usepackage{algorithm}
\usepackage{algpseudocode}

\newcommand{\NC}{\textsc{nc}}
\newcommand{\DAC}{\textsc{dac}}
\newcommand{\DDC}{\textsc{ddc}}
\newcommand{\TLC}{\textsc{tlc}}
\newcommand{\EPr}{\textsc{ep}}
\newcommand{\TTC}{\textsc{ttc}}
\newcommand{\LK}{\textsc{lk}}
\newcommand{\HC}{\textsc{hc}}
\newcommand{\EC}{\textsc{ec}}
\newcommand{\C}{\textsc{comfort}}
\newcommand{\EPDMS}{\mbox{EPDMS}}

\definecolor{wacvblue}{rgb}{0.21,0.49,0.74}
\usepackage[pagebackref,breaklinks,colorlinks,allcolors=wacvblue]{hyperref}

\def\wacvPaperID{1337}
\def\confName{WACV}
\def\confYear{2027}

\title{Designing Versatile Samples for Learned Trajectory Scoring}

\author{Yaguang Li\\
Purdue University\\
{\tt\small yaguangl@purdue.edu}
\and
Jiaru Zhang\\
Purdue University\\
{\tt\small jiaru@purdue.edu}
}

\author{Yaguang Li\textsuperscript{1}, Jiaru Zhang\textsuperscript{1}\thanks{Corresponding author: Jiaru Zhang.}, Chuheng Wei\textsuperscript{1}, Can Cui\textsuperscript{2}, Ziran Wang\textsuperscript{1}\\
Purdue University\textsuperscript{1}, Bosch Center for Artificial Intelligence\textsuperscript{2}\\
{\tt\small \{yaguangl, jiaru\}@purdue.edu
}
}

\begin{document}
\maketitle
\begin{abstract}
Many current end-to-end driving policies emit a pool of candidate trajectories and select one, which makes selection a separable component: a scorer can be retrained while the planner, its backbone, and its trajectory generator all stay frozen. However, many strong planners concentrate their proposals around safe mode, providing limited supervision near decision boundaries. In this work, we design a training dataset that provides more informative supervision for the scorer. In particular, we construct two generators that perturb the logged human trajectory along the two axes a vehicle can be displaced: laterally toward the drivable boundary and longitudinally toward a leading vehicle. The designed dataset produces more informative positive and negative samples than the base planner's proposal pool. We attach a transformer-based scorer to two frozen generative planners, DiffusionDrive and MeanFuser, and train it on the NAVSIM navtrain dataset. The results of the experiments show that we achieve 90.1 EPDMS on DiffusionDrive and 90.4 EPDMS on MeanFuser when using ResNet-34, with 0.4 and 0.3 EPDMS respectively, from the designed training dataset. The code is publicly available at \url{https://github.com/PurdueDigitalTwin/Versatile-Scoring}.
\end{abstract}
\section{Introduction}
\label{sec:intro}

\begin{figure}[t]
\centering
\includegraphics[width=\columnwidth]{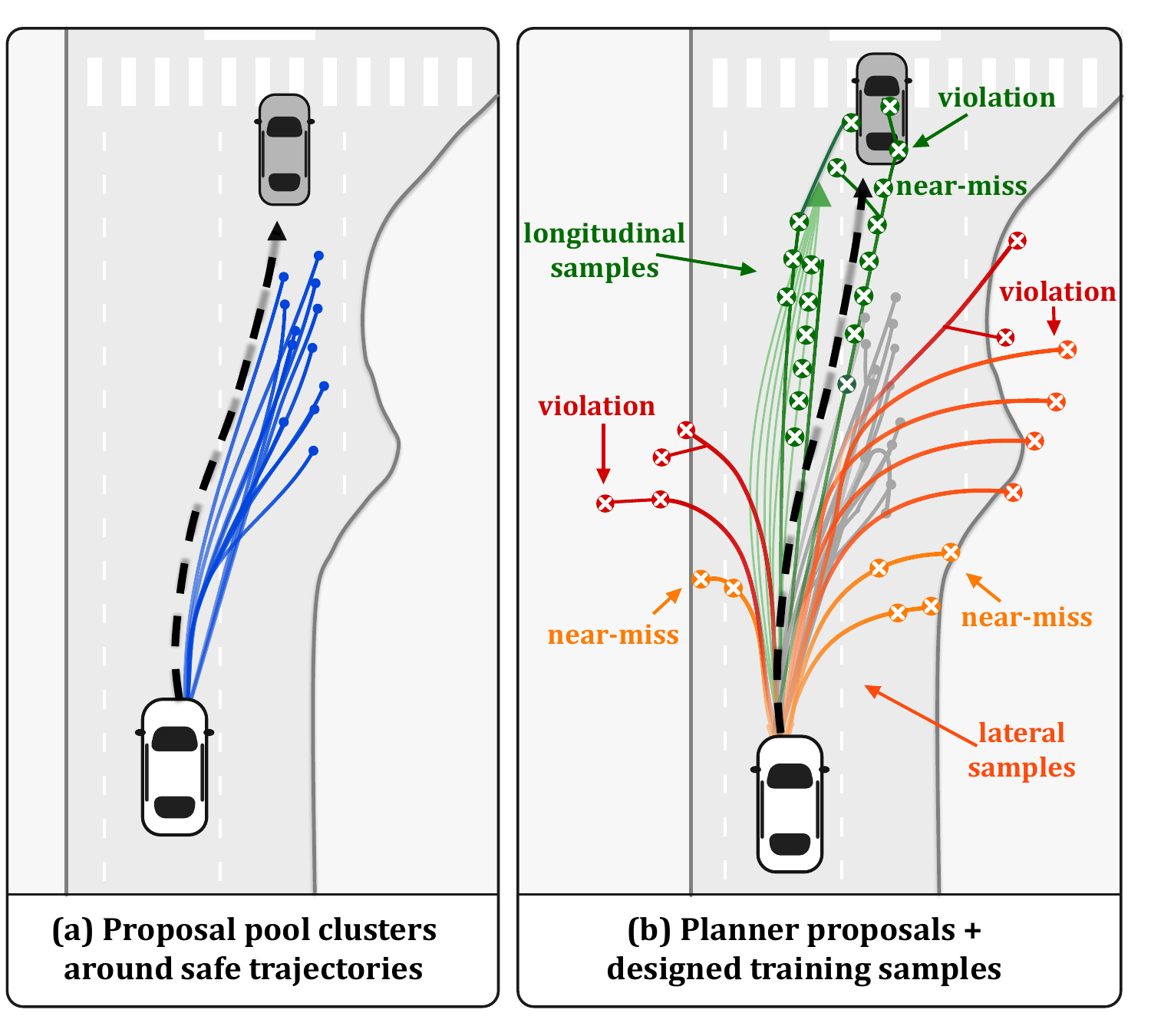}
\caption{Previous selection-based methods (a) are supervised only on the planner's own proposal pool (blue), whose candidates overlap heavily and are mostly collision-free. We augment each scene with synthetic trajectories (b) from lateral (orange) and longitudinal (green) displacement generators, which induce near-misses and true violations.}
\label{fig:teaser}
\end{figure}

Early end-to-end driving policies typically regress a single trajectory from sensor inputs~\cite{hu2023uniad,jiang2023vad}. However, a single output cannot express the multimodality of driving; therefore, recent policies instead emit a \emph{set} of candidates and select one from it~\cite{chitta2023transfuser,li2024hydramdp,liao2025diffusiondrive,meanfuser}. This splits the policy into two stages: a planner maps sensor inputs and ego status to a pool of candidate trajectories, and a scorer consumes the same scene context together with each candidate, ranks the pool, and emits one trajectory as the final output. The two stages are separable. A scorer can be retrained while the planner stays frozen, so the same scoring framework can be applied to different policies that expose candidate trajectories, at a fraction of the cost of retraining the entire policy. 

Decomposing the policy into a planner and a scorer opens two possible routes to better driving performance: building a powerful planner or a powerful scorer. Recent work \cite{liao2025diffusiondrive, meanfuser} has already produced strong generative planners, but comparable progress on the scorer has lagged. In fact, evidence from nuPlan \cite{caesar2021nuplan} shows a substantial gap between the trajectory a policy selects and the best trajectory in the pool that the same policy has already proposed~\cite{dauner2023parting}. Therefore, our objective is to close the gap in the scorer module.

Two main challenges remain to close the gap. The first challenge is that candidates near the decision boundaries are rare. A competent planner produces few boundary or failure cases, providing limited supervision for learning the boundary region. As illustrated in \cref{fig:teaser} (a), the base planner's proposals concentrate in a narrow region of trajectory space, which makes it difficult for the scoring module to select the best trajectory candidate from the proposal pool. 

The second challenge is that many components of driving performance are binary-labeled, such as collision with other objects and compliance with traffic rules. The components are also cascaded, meaning a single violation zeros the entire driving performance. Two trajectories separated by centimeters therefore receive opposite targets, and the scorer must resolve that discontinuity from nearly identical inputs. Previous work primarily addresses this issue by increasing the vocabulary of trajectory candidates. Hydra-MDP~\cite{li2024hydramdp} distills several rule-based teachers into a multi-head scorer over a fixed trajectory vocabulary. But the vocabulary in Hydra-MDP is fixed and scene-independent, which leaves it to chance whether any entry falls near this scene's decision boundary, so enlarging it mostly adds irrelevant candidates rather than informative ones. GTRS~\cite{li2025gtrs} pairs a diffusion-based proposal generator with a scorer trained on super-dense trajectory sets under dropout regularization. However, both offline labeling and the scoring pass scale with vocabulary size, putting an increasingly heavy burden on computational resources.

We address these limitations in two ways. First, we develop a transformer-based scorer that sits on top of frozen generative base planners, predicting each component of driving performance with independent decoders and composing them through the metric's own formula rather than regressing the scalar directly. The weights of the planner network stay frozen during the scorer's training, and one scorer trains in under an hour on a single GPU. Second, we design more data samples on top of the trajectory pool the base planner emits. Two generators are anchored on the recorded human trajectory and displace it along the two axes along which a vehicle can actually move: laterally toward the drivable-area boundary, and longitudinally along its own path toward a lead vehicle, as shown in \cref{fig:teaser}. The samples inside the boundary are near-misses that still pass, whereas those outside are violations. Our contributions are as follows:
\begin{itemize}
\item We develop a lightweight transformer-based scoring module on top of DiffusionDrive~\cite{liao2025diffusiondrive} and MeanFuser~\cite{meanfuser} and train it separately while the planning module remains frozen.
\item We design a versatile training dataset along the boundary region to study the scorer's training pool as a design variable and show that synthetic trajectories improve driving performance further.
\item We demonstrate consistent gains on the NAVSIM~\cite{dauner2024navsim} benchmark across two planners with different sensor configurations and pool sizes, and give a component-wise account showing that the gain concentrates in the multiplicative safety terms that the metric's structure makes decisive.
\end{itemize}
\section{Related Work}
\label{sec:related}

\subsection{End-to-End Autonomous Driving Planning}

Classical driving stacks separate perception, prediction and planning into independently trained modules, which incurs information loss and lets errors propagate across interfaces. UniAD~\cite{hu2023uniad} unified the stack into a single optimizable network, and a large body of end-to-end work has followed~\cite{jiang2023vad,chen2024vadv2,chitta2023transfuser,liao2025diffusiondrive,meanfuser,epona,sparsedrive}. Almost all of it is trained by imitation from logged expert demonstrations, since interaction with a simulator of sufficient fidelity remains expensive.

Early end-to-end policies regressed a single trajectory, either directly from sensors or through an explicit perception-prediction stack~\cite{hu2023uniad,jiang2023vad}. A single regressed output has two weaknesses. It cannot express the multi-modality of driving and it exposes no alternative against which the chosen behavior can be assessed. A policy can score well on open-loop benchmarks by extrapolating ego status alone~\cite{egostatus} and end-to-end models easily rely on shortcut features of the training distribution~\cite{hiddenbiases}. Recent policies therefore emit a set of candidates instead of one. VADv2 learns a distribution over a discrete planning vocabulary~\cite{chen2024vadv2}; Hydra-MDP ranks a fixed trajectory vocabulary~\cite{li2024hydramdp}; DiffusionDrive truncates a diffusion process into a handful of diverse modes~\cite{liao2025diffusiondrive}; GoalFlow conditions flow matching on predicted goal points~\cite{goalflow}; and MeanFuser generates its candidates in a single step from a mean-flow formulation~\cite{meanfuser}. Others condition generation on a learned world model~\cite{world4drive,epona,wote}. We evaluate on NAVSIM~\cite{dauner2024navsim}, which scores a policy in a non-reactive simulation with the extended PDM score (EPDMS), and whose scenario distribution is drawn from nuPlan~\cite{caesar2021nuplan,nuplanbench}.

\subsection{The Generate-Select Pipeline}

The generate-select pipeline shares a common structure: a generator produces a pool of candidate trajectories from the scene, and a scorer assigns each candidate a value. Hydra-MDP~\cite{li2024hydramdp} and Hydra-MDP++~\cite{hydramdppp} rank a vocabulary of clustered trajectories, and GTRS~\cite{li2025gtrs} pushes the vocabulary to an extreme density so that the decision boundary is populated by sheer volume, paired with a diffusion-based proposal generator and dropout regularization. Another line lets a generative planner produce a small, scene-conditioned pool (e.g., twenty modes for DiffusionDrive~\cite{liao2025diffusiondrive}, eight for MeanFuser~\cite{meanfuser}), which is far cheaper but leaves the scorer with very few candidates to learn from.

In terms of scoring, Hydra-MDP distills several rule-based teachers into a multi-head scorer, trading exactness for lower inference cost~\cite{li2024hydramdp}. DriveSuprim~\cite{drivesuprim} keeps the dense vocabulary but rescores it in two stages, filtering $8192$ candidates to $256$ before a refinement decoder separates them. ZTRS~\cite{li2025ztrs} removes imitation from the objective and trains a scorer directly on simulator returns. DrivoR~\cite{kirby2026drivor} compresses multi-camera ViT features into camera-aware register tokens that drive two lightweight decoders, one generating candidates and one scoring them. We develop our scoring module differently: our scorer predicts the nine \EPDMS{} components of \cref{eq:epdms} and is trained standalone over a planner that is frozen end to end, including its perception backbone, so no gradient from the scoring objective reaches any part of the generator and the same scorer design can be attached to planners it was not co-designed with.

\subsection{Dataset Augmentation for Driving}

Behavior cloning is limited by the states its demonstrator visits: an expert policy rarely approaches a rule boundary, so a model trained on its logs is never shown what approaching one looks like, and cannot recover from states the expert never entered~\cite{galashov2022data, ross2011dagger, behaviorcloning}. Fitting a value or reward model has the same requirement in sharper form, since a value function must see examples on both sides of a decision boundary to place it, and offline RL has long noted that a behavior policy's own data does not supply them~\cite{kumar2020cql, kostrikov2022iql, park2023hiql}.

Perturbing expert trajectories is the established response. ChauffeurNet~\cite{bansal2019chauffeurnet} deviates from the logged path and synthesizes collisions and off-road excursions so the policy learns to recover from them; DAgger-style aggregation collects corrective labels along the learner's own distribution~\cite{dataaggregation, cheating}; and KING~\cite{king} optimizes other agents' trajectories through a kinematics model to manufacture collisions. A parallel line in reinforcement learning generates counterfactual transitions from factored dynamics for the same reason~\cite{counterfactual}. Recent work scales trajectory perturbation with modern rendering: neural sensor simulators reconstruct a scene and re-render it from deviated poses~\cite{unisim, neurad, zhou2026hugsim}, and SimScale~\cite{tian2026simscale} perturbs the ego trajectory, renders the multi-view observations the ego would have seen at the deviated state, and co-trains the planner on real and simulated data.

\section{Preliminaries}
\label{sec:prelim}

\subsection{EPDMS Metric}
\label{sec:method:setting}

\begin{figure*}
\centering
\includegraphics[width=\textwidth]{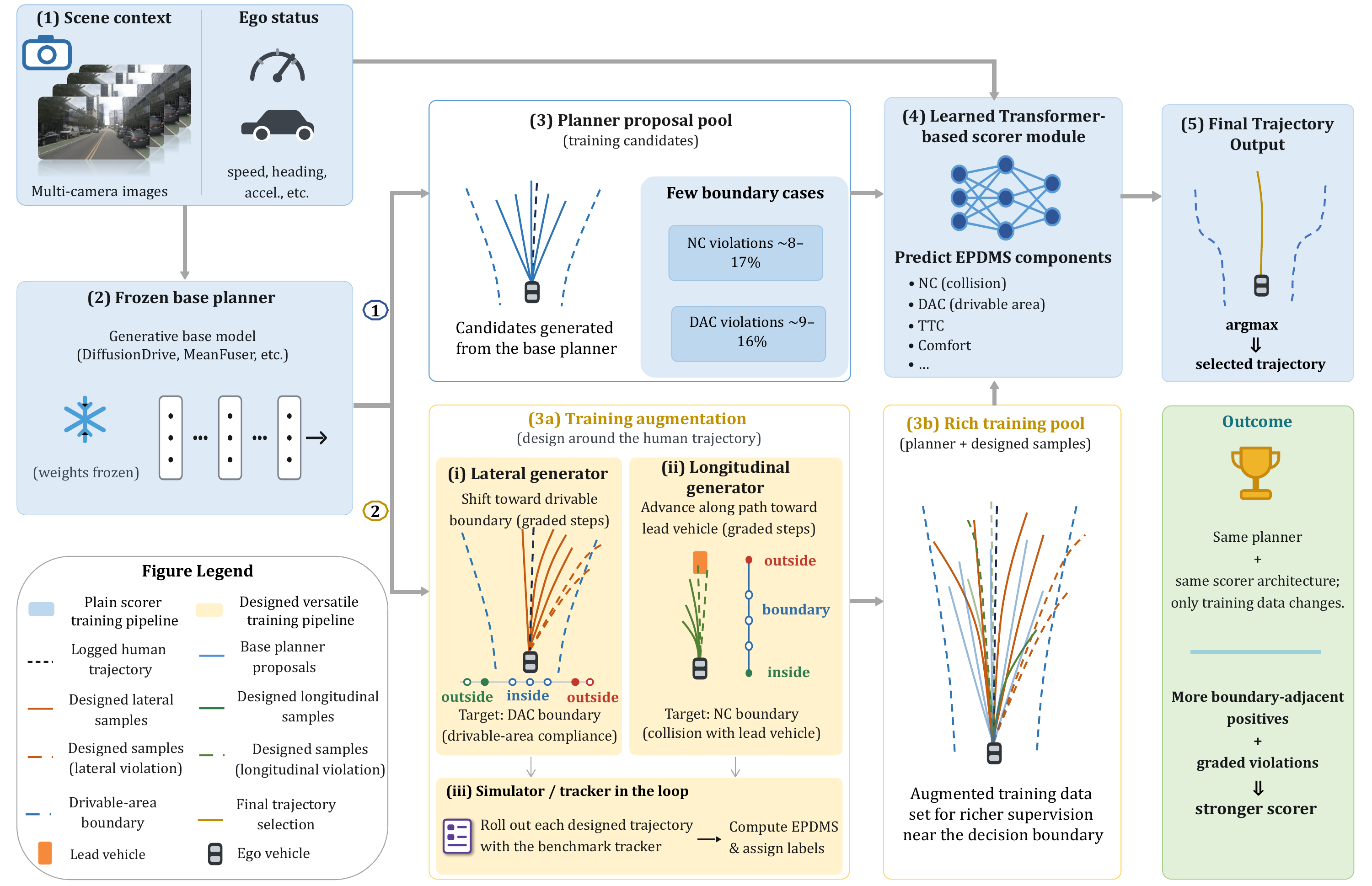}
\caption{\textbf{Overall architecture.} Sensor inputs and ego status drive a frozen generative planner, which emits a pool of candidate trajectories. A learned scorer consumes the same scene context together with each candidate, predicts the metric's components, composes them into an EPDMS estimate and deploys the argmax. What we change is the pool the scorer is \emph{trained} on (shown in orange), which at training time carries designed samples alongside the planner's own candidates, and at inference reverts to the planner's candidates alone.}
\label{fig:architecture}
\end{figure*}

On NAVSIM~\cite{dauner2024navsim}, a policy observes a scene and emits candidates $\{\tau_1,\dots,\tau_N\}$, each a sequence of $8$ waypoints $(x,y,\theta)$ over a $4$\,s horizon at $2$\,Hz. The benchmark scores a trajectory with the extended PDM score, a product of four rule terms and a weighted average of five quality terms:
\begin{equation}
\label{eq:epdms}
\begin{aligned}
\EPDMS = {} & \NC\cdot\DAC\cdot\DDC\cdot\TLC \\[2pt]
            & \times\ (5\EPr + 5\TTC + 2\LK + 2\HC + 2\EC)/16.
\end{aligned}
\end{equation}

The multiplicative block is the reason \NC{} and \DAC{} dominate: a single violation zeros the score regardless of comfort or progress.

\subsection{System Architecture}
\label{sec:method:scorer}

\Cref{fig:architecture} shows the overall architecture. A planner maps sensor inputs and ego status to a pool of candidate trajectories. On navtrain, $8.4\%$ of MeanFuser's~\cite{meanfuser} proposals and $16.6\%$ of DiffusionDrive's~\cite{liao2025diffusiondrive} proposals fail NC, and $8.9\%$ of MeanFuser's~\cite{meanfuser} proposals and $15.8\%$ of DiffusionDrive's~\cite{liao2025diffusiondrive} proposals fail DAC. A scorer then ranks that pool and emits one trajectory as the final output. The two stages are separable, which means the scorer can be retrained without touching the planner, its backbone, or its candidate generator, and any improvement transfers to every policy that exposes its candidates. Note that the scorer receives the scene context directly rather than through the planner's decision, so it is free to disagree with the planner's own ranking.

\paragraph{Scorer Module.}
We score candidates by predicting the metric's components separately with small transformer decoders, rather than regressing the scalar directly. Our module consists of nine independent transformer networks that predict components of EPDMS in \cref{eq:epdms} and is trained standalone over a frozen planner.

We train our scoring network based on each of the nine EPDMS components, sharing no weights. Each network takes two inputs: the frozen planner's scene context (a short sequence of feature tokens of width $128$ for the MeanFuser~\cite{meanfuser} and $256$ for DiffusionDrive~\cite{liao2025diffusiondrive}) and a candidate trajectory. The trajectory is flattened and lifted to a single query by a two-layer MLP; the queries of a scene pass through a three-layer transformer decoder that cross-attends to the context; and a two-layer MLP head reduces each decoded query to one number.

\paragraph{Training Objective.}
Let $\hat{y}_j$ be the predicted value of component $j$, and let $y_j$ be the simulator's label. NAVSIM leaves a component unlabeled when the rule does not apply to the particular scene, so not every component is labeled for every candidate. Let $\mathcal{C}_j$ denote the candidates in the batch whose $j$-th label is defined. The eight binary components use a focal BCE~\cite{lin2017focal} that emphasizes the rare failures,
\begin{equation}
\label{eq:focal}
\ell^{\text{foc}}_j = -\,(1-p_j)^{\gamma}\,\log p_j,
\qquad
p_j = \hat{y}_j y_j + (1-\hat{y}_j)(1-y_j),
\end{equation}
with $\gamma = 2$, while ego progress is a continuous ratio and uses a squared
error. The scoring loss is their masked mean, summed over components with weights
$\lambda_j$:
\begin{equation}
\label{eq:loss}
\mathcal{L} = \sum_{j \neq \EPr} \frac{\lambda_j}{|\mathcal{C}_j|}
              \sum_{\mathcal{C}_j} \ell^{\text{foc}}_j
            + \frac{\lambda_{\EPr}}{|\mathcal{C}_{\EPr}|}
              \sum_{\mathcal{C}_{\EPr}} (\hat{y}_{\EPr} - y_{\EPr})^2 .
\end{equation}
We set every $\lambda_j = 1$; the networks are independent, so no re-weighting was needed to keep them from competing. The undefined label produces no gradients to the network.

\paragraph{Deployment.}
At inference, the nine predictions for a candidate are composed into a single score through \cref{eq:epdms}, and the candidate with the highest composed EPDMS is driven. Checkpoints are selected on the best \emph{realized} EPDMS over a held-out validation split of whole logs.

\section{Designing Versatile Trajectory Samples}
\label{sec:method}

Let $\tau_h$ be the logged human trajectory of a scene and $\{\tau_i\}_{i=1}^{N}$ be the $N$ candidates the frozen planner proposes for it. From these we build $K$ \emph{rungs}: perturbed copies of $\tau_h$ whose margins sit at prescribed distances from violating a rule.

\Cref{fig:design} shows the designs of our trajectories: each is anchored on the human trajectory and walks toward one rule in graded steps, so the ladder straddles the boundary rather than landing on one side of it. The two designs differ only in the direction of that walk, which is what makes them share the construction below. We build the training set in five steps:

\begin{figure}
\centering
\includegraphics[width=\columnwidth]{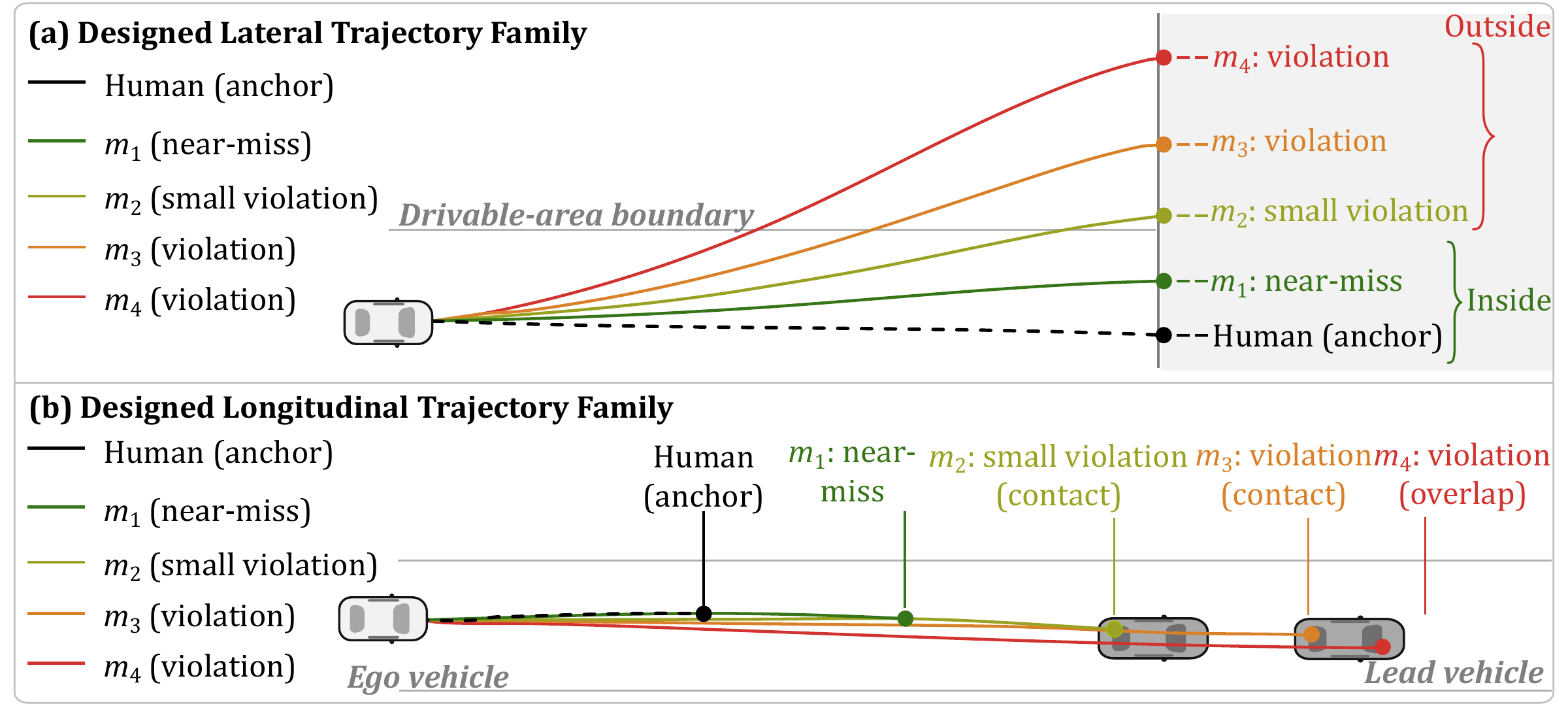}
\caption{The designs of the trajectory samples. \textbf{(a)} Our lateral design displaces the human path toward the nearest boundary in graded steps. \textbf{(b)} Our longitudinal design advances the ego along its own path toward the lead vehicle.}
\label{fig:design}
\end{figure}

\begin{enumerate}\itemsep1pt
\item \textbf{Measure} the human's own margin $M(\tau_h)$ under the rule in question (\cref{sec:method:margin}).
\item \textbf{Weight} each waypoint by a ramp $w(s_i)$ fixed by the human path alone (\cref{sec:method:ramp}).
\item \textbf{Build} a one-parameter family $\{\tau_\delta\}$ by displacing $\tau_h$ laterally and longitudinally (\cref{sec:method:family}).
\item \textbf{Set} $K$ target margins $m_1 > \dots > m_K$, corresponding to one near-miss and $K{-}1$ graded violations (\cref{sec:method:targets}).
\item \textbf{Solve} for the smallest $\delta$ reaching each target, with the simulator inside the loop (\cref{sec:method:executed}).
\end{enumerate}

The human margin $M(\tau_h)\in\mathbb{R}$ is a signed scalar such that $M < 0$ means violation, and $\delta$ is the displacement from the human baseline.

\subsection{Measure the Margin}
\label{sec:method:margin}

\paragraph{Lateral.}
$M_{\mathrm{lat}}(\tau)$ is the signed distance from the \emph{swept footprint} of $\tau$ to the drivable-area boundary (positive inside), evaluated on all four corners at all eight waypoints rather than on the centerline.

\paragraph{Longitudinal.}
$M_{\mathrm{lon}}(\tau)$ is the minimum gap over the horizon from the ego's \emph{front bumper line} to a chosen lead vehicle, counting only steps where that vehicle is ahead. 

\subsection{Weight the Ramp}
\label{sec:method:ramp}

All perturbations use a polynomial smooth step ramp over arc length $s$, zero at the
ego pose:
\begin{equation}
\label{eq:ramp}
w(s) = \bar{s}^2(3 - 2\bar{s}), \qquad \bar{s} = \mathrm{clip}(s/s_0,\,0,\,1),
\end{equation}
where the ramp scale is $s_0 = 10$\,m. Writing $s_i$ for the arc length of waypoint $i$ along $\tau_h$, both designs displace waypoint $i$ by the fraction $w(s_i)$ of $\delta$. The eight values $w(s_1),\dots,w(s_8)$ are therefore fixed once the human path is known, and are shared by both geometries.

\subsection{Build the Perturbation Family}
\label{sec:method:family}

\paragraph{Lateral.}
A lateral offset along the unit normal $n$ pointing away from the nearest boundary, with $n$ computed once per scene at the point of closest approach:
\begin{equation}
\label{eq:lat}
\tau_\delta[i] = \tau_h[i] + \delta\, w(s_i)\, n,
\end{equation}
where $\tau_h[i]$ is waypoint $i$ of the human trajectory, so a negative $\delta$ carries the ego toward the boundary. Because $w$ grows along the path, the rung bends into a lane change rather than sliding sideways; headings are recomputed from the offset path, not copied from $\tau_h$. 

\paragraph{Longitudinal.}
The ego advances \emph{along its own path} rather than sideways:
\begin{equation}
\label{eq:lon}
\tau_\delta[i] = \tau_h\big(s_i + \delta\, w(s_i)\big), \qquad \delta \ge 0,
\end{equation}
where $\tau_h(\cdot)$ reads the human path at a given arc length: the station advances, the path itself does not. While a rung stays on the logged path, this leaves \DAC{}, \LK{} and \DDC{} unchanged by construction; past its end the ego continues along its final heading.

\paragraph{Staying inside the planner's envelope.}
Both families are capped so that a rung is never longer than a trajectory the planner itself proposes: with $\{\tau_i\}$ the planner's $N$ proposals, the headroom is $R = \max_i \mathrm{len}(\tau_i) - \mathrm{len}(\tau_h)$, and scenes with $R < 0.5$\,m emit masked rungs. 

\subsection{Set the Target Schedule}
\label{sec:method:targets}

\paragraph{Lateral.}The lateral targets are set relative to the human's own clearance $m_h \equiv M_{\mathrm{lat}}(\tau_h)$:
\begin{equation}
\label{eq:lattargets}
\begin{aligned}
m_1 &= +\min(0.30\,m_h,\; 0.80), \\
m_{k+1} &= -\mathrm{clip}(\alpha_k m_h,\; \ell_k,\; u_k), \quad k = 1,2,3,
\end{aligned}
\end{equation}
with $\alpha = (0.10, 0.25, 0.50)$, $\ell = (0.15, 0.35, 0.60)$ and $u = (0.40, 0.80, 1.40)$ meters. Fixed percentages would give $7$\,cm excursions in tight scenes, which the tracker erases, and $>2$\,m excursions in open ones, which are trivially separable. The floors keep a target far enough out to survive tracking; the caps keep it close enough to hug the boundary.

\paragraph{Longitudinal.}The longitudinal targets are set at $(+0.5, -0.2, -0.6, -1.0)$\,m of bumper gap: one near-miss and three graded contacts.

\subsection{Solve for Displacement}
\label{sec:method:executed}

\begin{algorithm}[tb]
\small
\caption{Rung solving for one scene}
\label{alg:exec}
\begin{algorithmic}[1]
\Require human trajectory $\tau_h$, proposals $\{\tau_i\}$, targets $m_{1:K}$, margin $M$, family $\delta \mapsto \tau_\delta$, evaluator $E$ 
\Ensure $K$ rungs $\{\tau_{\delta_1},\dots,\tau_{\delta_K}\}$, each emitted or masked
\State $R \gets \max_i \mathrm{len}(\tau_i) - \mathrm{len}(\tau_h)$ \Comment{on-manifold headroom}
\If{$R < 0.5$\,m} \Return $K$ masked rungs \EndIf
\State $\Delta \gets N_\Delta$ magnitudes spanning the design's range, capped by $\min(\delta_{\max}, R)$
\State $\{\hat{\tau}_\delta\}_{\delta\in\Delta} \gets E(\{\tau_\delta\})$ \Comment{states to score: $E = \mathrm{sim}$ (batched) or identity}
\State $m_\delta \gets M(\hat{\tau}_\delta)$ for all $\delta \in \Delta$
\For{$k = 1 \dots K$}
  \State $\mathcal{H} \gets \{\delta \in \Delta : m_\delta \le m_k\}$ \Comment{magnitudes reaching target $k$}
  \If{$\mathcal{H} = \emptyset$} mask rung $k$
  \Else
    \State $\delta_k \gets \arg\min_{\delta \in \mathcal{H}} |\delta|$
    \State refine $\delta_k$ by bisection between $\delta_k$ and its neighbour in $\Delta$ that misses $m_k$
    \State emit $\tau_{\delta_k}$
  \EndIf
\EndFor
\end{algorithmic}
\end{algorithm}

After the targets are set, we recover the smallest $\delta$ in displacement space that reaches each target. It is recovered by inverting the map $\delta \mapsto M(\tau_\delta)$, taking the smallest displacement that reaches the target so the rung stays as close to the boundary as it can:
\begin{equation}
  \delta_k = \min \bigl\{\, \delta \in \Delta \;:\; M(\tau_\delta) \le m_k \,\bigr\},
  \label{eq:solve}
\end{equation}
where $\Delta$ is a grid of $N_\Delta$ trial magnitudes spanning the design's admissible range. The inversion is numerical: evaluate $M(\tau_\delta)$ over a grid of trial displacements, find where the resulting curve crosses $m_k$, and take that crossing. One $\delta$ is solved per rung, and each of $K$ rungs is either emitted or masked.

The perturbations are solved using simulation: NAVSIM does not apply its evaluation metrics to waypoints. It converts a trajectory to a reference and tracks that reference with a batch LQR controller driving a kinematic bicycle model. The candidate trajectories are measured on the rollout rather than on the waypoints, so there is a difference between the designed waypoints and the trajectory actually tracked by the simulator.
\section{Experiments}
\label{sec:exp}

\begin{table*}
\centering
\small
\setlength{\tabcolsep}{5pt}
\renewcommand{\arraystretch}{1.15}
\begin{tabular}{l|c|c|c|c c c c c|c}
\toprule
Method & Venue & Input & Backbone & \NC$\uparrow$ & \DAC$\uparrow$ & \TTC$\uparrow$ & \C$\uparrow$ & \EPr$\uparrow$ & PDMS$\uparrow$ \\
\midrule
TransFuser~\cite{chitta2023transfuser}       & IEEE TPAMI & C \& L & ResNet-34   & 97.7 & 92.8 & 92.8 & \textbf{100} & 79.2 & 84.0 \\
VADv2~\cite{chen2024vadv2}                 & arXiv 2024 & C \& L & ResNet-34   & 97.2 & 89.1 & 91.6 & \textbf{100} & 76.0 & 80.9 \\
Hydra-MDP~\cite{li2024hydramdp}             & arXiv 2024 & C \& L & ResNet-34   & 98.3 & 96.0 & 94.6 & \textbf{100} & 78.7 & 86.5 \\
GoalFlow~\cite{goalflow}          & CVPR 2025  & C \& L & ResNet-34   & 98.3 & 93.8 & 94.3 & \textbf{100} & 79.8 & 85.7 \\
DiffusionDrive~\cite{liao2025diffusiondrive} & CVPR 2025 & C \& L & ResNet-34  & 98.2 & 96.2 & 94.7 & \textbf{100} & 82.2 & 88.1 \\
WoTE~\cite{wote}                   & ICCV 2025  & C \& L & ResNet-34   & 98.5 & 96.8 & 94.9 & 99.9 & 81.9 & 88.3 \\
UniAD~\cite{hu2023uniad}                 & CVPR 2023  & C & ResNet-34        & 97.8 & 91.9 & 92.9 & \textbf{100} & 78.8 & 83.4 \\
World4Drive~\cite{world4drive}     & ICCV 2025  & C & ResNet-34        & 97.4 & 94.3 & 92.8 & \textbf{100} & 79.9 & 85.1 \\
MeanFuser~\cite{meanfuser}         & CVPR 2026  & C & ResNet-34        & \textbf{98.6} & 97.0 & \textbf{95.0} & \textbf{100} & 82.8 & 89.0 \\
\midrule
DiffusionDrive $+$ plain scorer    & -- & C \& L & ResNet-34 & 98.4 & 97.5 & 94.2 & \textbf{100} & 84.1 & 89.3 \\
DiffusionDrive $+$ augmented scorer   & -- & C \& L & ResNet-34 & \textbf{98.6} & 97.7 & 94.3 & \textbf{100} & 84.2 & 89.5 \\
MeanFuser $+$ plain scorer         & -- & C & ResNet-34      & 98.2 & 98.0 & 93.8 & \textbf{100} & 84.8 & 89.7 \\
\textbf{MeanFuser $+$ augmented scorer}        & -- & C & ResNet-34      & 98.3 & \textbf{98.3} & 93.7 & \textbf{100} & \textbf{85.1} & \textbf{89.9} \\
\bottomrule
\end{tabular}
\caption{\textbf{Performance on the NAVSIM-v1 \cite{dauner2024navsim} navtest benchmark.} ``C'' denotes camera, ``L'' LiDAR. All rows share the ResNet-34 \cite{he2016resnet} image backbone, so the comparison is at equal perception capacity. Our checkpoints are selected on the held-out validation split and every row of ours is the mean over three seeds. The baseline results are taken from~\cite{meanfuser}.}
\label{tab:navsim_v1}
\end{table*}

\begin{table*}[t]
\centering
\small
\setlength{\tabcolsep}{5pt}
\renewcommand{\arraystretch}{1.15}
\begin{tabular}{l|c|c c c c c c c c c|c}
\toprule
Method & Backbone & \NC$\uparrow$ & \DAC$\uparrow$ & \DDC$\uparrow$ & \TLC$\uparrow$ & \EPr$\uparrow$ & \TTC$\uparrow$ & \LK$\uparrow$ & \HC$\uparrow$ & \EC$\uparrow$ & EPDMS$\uparrow$ \\
\midrule
Ego Status MLP~\cite{dauner2024navsim} & --        & 93.1 & 77.9 & 92.7 & 99.6 & 86.0 & 91.5 & 89.4 & \textbf{98.3} & 85.4 & 64.0 \\
TransFuser~\cite{chitta2023transfuser}         & ResNet-34 & 96.9 & 89.9 & 97.8 & 99.7 & 87.1 & 95.4 & 92.7 & \textbf{98.3} & 87.2 & 76.7 \\
Hydra-MDP++~\cite{hydramdppp}        & ResNet-34 & 97.2 & 97.5 & 99.4 & 99.6 & 83.1 & 96.5 & 94.4 & 98.2 & 70.9 & 81.4 \\
GTRS-Dense$^\dagger$~\cite{li2025gtrs} & ResNet-34 & 97.6 & 97.5 & 99.0 & \textbf{99.9} & 87.9 & 97.0 & 95.9 & 97.5 & 55.9 & 82.3 \\
DriveSuprim~\cite{drivesuprim}       & ResNet-34 & 97.5 & 96.5 & 99.4 & 99.6 & 88.4 & 96.6 & 95.5 & \textbf{98.3} & 77.0 & 83.1 \\
DiffusionDrive~\cite{liao2025diffusiondrive} & ResNet-34 & 98.2 & 96.3 & 99.4 & 99.8 & 87.4 & 97.4 & 97.0 & \textbf{98.3} & 87.7 & 88.3 \\
MeanFuser~\cite{meanfuser}           & ResNet-34 & 98.3 & 97.2 & \textbf{99.6} & 99.8 & 87.6 & 97.4 & \textbf{97.3} & \textbf{98.3} & \textbf{88.2} & 89.5 \\
\midrule
DiffusionDrive $+$ plain scorer      & ResNet-34 & 98.4 & 97.5 & 99.5 & 99.8 & 88.2 & 97.6 & 96.3 & \textbf{98.3} & 87.7 & 89.7 \\
DiffusionDrive $+$ augmented scorer  & ResNet-34 & \textbf{98.6} & 97.7 & 99.5 & \textbf{99.9} & 88.0 & \textbf{97.9} & 96.2 & \textbf{98.3} & 87.5 & 90.1 \\
MeanFuser $+$ plain scorer           & ResNet-34 & 98.2 & 98.0 & \textbf{99.6} & 99.8 & \textbf{88.7} & 97.5 & 96.8 & \textbf{98.3} & 87.0 & 90.1 \\
\textbf{MeanFuser $+$ augmented scorer}          & ResNet-34 & 98.3 & \textbf{98.3} & \textbf{99.6} & 99.8 & 88.6 & 97.4 & 96.6 & \textbf{98.3} & 87.2 & \textbf{90.4} \\
\bottomrule
\end{tabular}
\caption{\textbf{Performance on the NAVSIM-v2~\cite{cao2025navsim_v2} navtest benchmark.} Every entry except the sensor-free Ego Status MLP uses a ResNet-34~\cite{he2016resnet} image backbone. Our checkpoints are selected on the held-out validation split, and every row of ours is the mean over three seeds. The baseline results are taken from~\cite{meanfuser}. $\dagger$ Result is reported by~\cite{tian2026simscale}.}
\label{tab:navsim_v2}
\end{table*}

\subsection{Setup}
\label{sec:exp:protocol}

We evaluate on the NAVSIM navtest split ($12{,}146$ scenes) over two frozen planners chosen for their contrasting candidate pools: DiffusionDrive~\cite{liao2025diffusiondrive}, a camera and LiDAR planner emitting $N=20$ proposals, and MeanFuser~\cite{meanfuser}, a camera-only planner emitting $N=8$. Both planners are held fixed throughout the training process and the transformer scorers are trained separately and attached to the base planners. We use our own designed augmented dataset to train the scorer instead of directly adopting the output trajectories of the base planners.

Each configuration is trained for $100$ epochs. Training one scorer takes approximately $50$ minutes on a single RTX 5090 and the deployed checkpoint is the epoch with the best EPDMS on the held-out validation split, which accounts for approximately 5\% of the entire NAVSIM training dataset.

\subsection{NAVSIM Navtest Result}
\label{sec:exp:main}

\Cref{tab:navsim_v1,tab:navsim_v2} compare our approach with published NAVSIM entries under both metric versions. All the baseline methods and our base planners use the same ResNet-34 backbone for training and evaluation. The baseline numbers are quoted from~\cite{meanfuser}, except where a caption notes a different source. Re-ranking a frozen planner's own proposals with a scorer trained on the augmented dataset raises DiffusionDrive from $88.1$ to $89.5$ PDMS and from $88.3$ to $90.1$ EPDMS, and MeanFuser from $89.0$ to $89.9$ PDMS and from $89.5$ to $90.4$ EPDMS. Neither planner, backbone, nor candidate generator is modified. 

Both tables also list the same scorer trained on the planner's own proposals alone, so the two sources of improvement can therefore be separated directly. The gain over the \emph{base planner} is large: up to $1.8$ EPDMS on DiffusionDrive~\cite{liao2025diffusiondrive} and the plain scorer accounts for 1.4, since the planner's own pick is far from the best candidate it already proposed. The gain attributable to \emph{our augmentation}, measured against the base scorer trained on the planner's proposals alone, is $0.4$ EPDMS on DiffusionDrive and $0.3$ EPDMS on MeanFuser.

The sub-scores also show what the gain costs. Drivable-area compliance rises on both planners and most of all on MeanFuser, while ego progress and lane keeping fall slightly on both. The scorer is not uniformly better, which \cref{sec:exp:headroom} examines in more detail.

\begin{figure}
\centering
\includegraphics[width=\columnwidth]{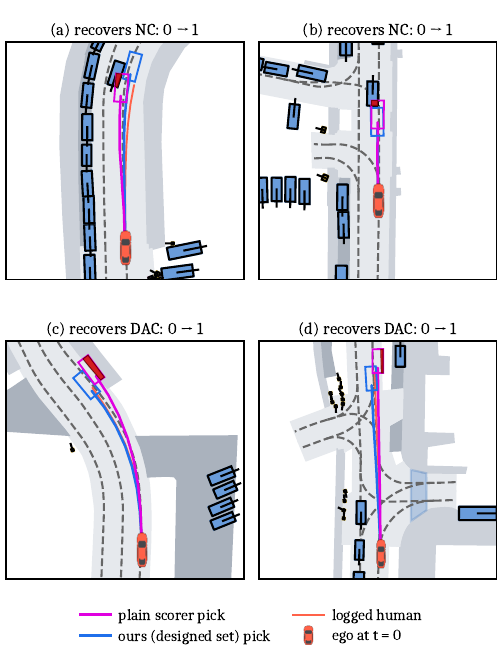}
\caption{Recovered gates on MeanFuser~\cite{meanfuser}. Four navtest scenes in which the two scorers deploy different candidates from the \emph{same} frozen $8$-proposal pool: the plain scorer's pick (magenta) is scored $0$ on a multiplicative gate and ours (blue) is scored $1$.}
\label{fig:qualitative}
\end{figure}

\Cref{fig:qualitative} shows what that improvement looks like on individual scenes. Over all $12{,}146$ scenes, the designed set cuts the rate at which a gate-failing candidate is deployed from $3.70\%$ to $3.35\%$ (\DAC{} $2.05\%\!\rightarrow\!1.73\%$, \NC{} $1.70\%\!\rightarrow\!1.65\%$).

\subsection{Ablation Study}
\paragraph{Component Score Analysis.}
\label{sec:exp:headroom}

\begin{table*}
\centering
\small
\setlength{\tabcolsep}{7pt}
\renewcommand{\arraystretch}{1.12}
\begin{tabular}{lcccc|cccc}
\toprule
& \multicolumn{4}{c|}{DiffusionDrive~\cite{liao2025diffusiondrive}} & \multicolumn{4}{c}{MeanFuser~\cite{meanfuser}} \\
Component & plain & our design & gap closed & ceiling & plain & our design & gap closed & ceiling \\
\midrule
\NC   & 98.39 & \textbf{98.56} & $+12.4$\% & 99.76 & 98.22 & \textbf{98.27} & \ $+4.8$\% & 99.26 \\
\DAC  & 97.52 & \textbf{97.70} & $+10.0$\% & 99.32 & 97.95 & \textbf{98.27} & $+26.7$\% & 99.15 \\
\DDC  & 99.50 & \textbf{99.54} & $+10.5$\% & 99.88 & 99.61 & \textbf{99.61} & \ $+0.0$\% & 99.84 \\
\TLC  & 99.83 & \textbf{99.88} & $+29.4$\% & 100.00 & 99.81 & \textbf{99.83} & $+25.0$\% & 99.89 \\
\EPr  & 88.15 & 87.96 & \ $-1.9$\% & 98.15 & 88.68 & 88.65 & \ $-0.6$\% & 94.13 \\
\TTC  & 97.63 & \textbf{97.87} & $+11.8$\% & 99.66 & 97.45 & 97.43 & \ $-1.6$\% & 98.74 \\
\LK   & 96.33 & 96.20 & \ $-4.0$\% & 99.54 & 96.77 & 96.56 & \ $-9.1$\% & 99.07 \\
\HC   & 98.33 & 98.32 & \ $-5.3$\% & 98.52 & 98.28 & \textbf{98.28} & \ $+0.0$\% & 98.36 \\
\EC   & 87.70 & 87.51 & \ $-2.2$\% & 96.38 & 87.05 & \textbf{87.22} & \ $+2.1$\% & 95.05 \\
\midrule
EPDMS & 89.71 & \textbf{90.05} & \ $+5.52$\% & 95.87 & 90.12 & \textbf{90.41} & \ $+6.89$\% & 94.33 \\
\bottomrule
\end{tabular}
\caption{Realised component scores on DiffusionDrive \cite{liao2025diffusiondrive} and MeanFuser \cite{meanfuser} over the deployed pick, averaged over three seeds. Ceiling indicates best-of-N taken independently across N proposals. \% of gap closed is $(\text{ours}-\text{plain})/(\text{ceiling}-\text{plain})$, the fraction of the available headroom the augmentation recovers. Every term that gates the product improves or holds on both planners, and what pays for them is ego progress, lane keeping, and comfort, with EPDMS closing $5.52\%$ and $6.89\%$ of the gap to the best-of-N ceiling.}
\label{tab:headroom}
\end{table*}

\Cref{tab:headroom} reports EPDMS component under the deployed pick, together with its oracle ceiling (best-of-N) and the fraction of that gap the design closes. The split follows the metric's own structure on both planners. Every safety gate improves or holds: all four close $10.0$--$29.4\%$ of their headroom on DiffusionDrive, and on MeanFuser three close $4.8$--$26.7\%$ while driving-direction compliance (\DDC) is unchanged. Lane keeping (\LK) and ego progress (\EPr) are worse, at $-4.0\%$ and $-1.9\%$ on DiffusionDrive and $-9.1\%$ and $-0.6\%$ on MeanFuser, and time-to-collision (\TTC) adds $11.8\%$ on DiffusionDrive but is flat on MeanFuser. History comfort (\HC) starts close to its ceiling, so the change is small. The overall EPDMS gain is positive on both: $5.52\%$ of the distance to the oracle on DiffusionDrive and $6.89\%$ on MeanFuser, which shows the effectiveness of our designed training dataset.

\paragraph{Designed Against Random Samples.}
\label{sec:exp:random}

In order to identify whether the augmentation of the dataset or the design itself contributes to the overall improvement, we pair the designed set with a randomly sampled set matched for the number of supervised rows and delivered to the same heads in the same numbers. \cref{tab:random} shows that the random samples contribute very little to the performance improvement. On DiffusionDrive it moves \DAC{} only from $97.52$ to $97.54$, and on MeanFuser only from $97.95$ to $97.98$, while the design reaches $97.70$ and $98.27$. The resulting EPDMS values are: $89.72$ and $90.22$ for the random, against $90.05$ and $90.41$ for the design.

\begin{table}
\centering
\small
\setlength{\tabcolsep}{4pt}
\begin{tabular}{lccc}
\toprule
 & Plain scorer & Random samples & \textbf{Design samples} \\
\midrule
\multicolumn{4}{l}{\emph{DiffusionDrive}~\cite{liao2025diffusiondrive}} \\
EPDMS  & $89.71_{\pm.03}$ & $89.72_{\pm.06}$ & $\mathbf{90.05}_{\pm.01}$ \\
\NC{}  & $98.39_{\pm.03}$ & $98.37_{\pm.06}$ & $\mathbf{98.56}_{\pm.05}$ \\
\DAC{} & $97.52_{\pm.02}$ & $97.54_{\pm.09}$ & $\mathbf{97.70}_{\pm.05}$ \\
\midrule
\multicolumn{4}{l}{\emph{MeanFuser}~\cite{meanfuser}} \\
EPDMS  & $90.12_{\pm.01}$ & $90.22_{\pm.02}$ & $\mathbf{90.41}_{\pm.05}$ \\
\NC{}  & $98.22_{\pm.05}$ & $98.24_{\pm.01}$ & $\mathbf{98.27}_{\pm.07}$ \\
\DAC{} & $97.95_{\pm.06}$ & $97.98_{\pm.10}$ & $\mathbf{98.27}_{\pm.03}$ \\
\bottomrule
\end{tabular}
\caption{Designed dataset against its random samples. We do an ablation on random samples with the same amount of data added to the plain scorer dataset ($437$k). Every entry is the mean $\pm$ standard deviation over three seeds. The results on DiffusionDrive \cite{liao2025diffusiondrive} and MeanFuser \cite{meanfuser} indicate that the majority of the improvement is due to the design itself rather than plain augmentation.}
\label{tab:random}
\end{table}

\paragraph{Decomposition of the Longitudinal and Lateral Designs.}

We study the contribution for each of the designs, and summarize in \cref{tab:generators}. 
The results show that either design alone already improves on the plain scorer: $89.96$ and $89.89$ against $89.71$ on DiffusionDrive, $90.24$ and $90.36$ against $90.12$ on MeanFuser. With both designs, the score is even better, at $90.05$ and $90.41$. Neither dominates, and each shows up in the component it supervises: the lateral design alone already recovers the full \DAC{} gain, $97.72$ and $98.26$ against the full design's $97.70$ and $98.27$, while the longitudinal design alone does the same for \NC{}, $98.54$ and $98.28$ against $98.56$ and $98.27$. Each generator contributes to the gain of the corresponding component, and their combination attains the best overall driving performance.

\begin{table}
\centering
\small
\setlength{\tabcolsep}{4pt}
\begin{tabular}{lcccc}
\toprule
 & Plain & Lon. only & Lat. only & \textbf{Both} \\
\midrule
\multicolumn{5}{l}{\emph{DiffusionDrive}~\cite{liao2025diffusiondrive}} \\
EPDMS  & $89.71_{\pm.03}$ & $89.96_{\pm.09}$ & $89.89_{\pm.09}$ & $\mathbf{90.05}_{\pm.01}$ \\
\NC{}  & $98.39_{\pm.03}$ & $98.54_{\pm.09}$ & $98.47_{\pm.09}$ & $\mathbf{98.56}_{\pm.05}$ \\
\DAC{} & $97.52_{\pm.02}$ & $97.62_{\pm.08}$ & $\mathbf{97.72}_{\pm.04}$ & $97.70_{\pm.05}$ \\
\midrule
\multicolumn{5}{l}{\emph{MeanFuser}~\cite{meanfuser}} \\
EPDMS  & $90.12_{\pm.01}$ & $90.24_{\pm.08}$ & $90.36_{\pm.04}$ & $\mathbf{90.41}_{\pm.05}$ \\
\NC{}  & $98.22_{\pm.05}$ & $\mathbf{98.28}_{\pm.02}$ & $98.26_{\pm.03}$ & $98.27_{\pm.07}$ \\
\DAC{} & $97.95_{\pm.06}$ & $98.01_{\pm.12}$ & $98.26_{\pm.01}$ & $\mathbf{98.27}_{\pm.03}$ \\
\bottomrule
\end{tabular}
\caption{Contribution of each generator, reported as mean $\pm$ std over three seeds. Each column keeps one generator and drops the other; the last column is the full designed set. Either design alone already recovers its own component's gain on both planners.}
\label{tab:generators}
\end{table}
\section{Conclusion}
\label{sec:conclusion}

We present a designed training set for learned trajectory scoring, built by perturbing the logged human trajectory laterally and longitudinally in steps graded against each scene's own geometry and labeled through the benchmark's own tracker. Our results highlight that the pool on which a scorer is trained is a design variable in its own right: a frozen planner, a transformer-based scorer, and a redesigned training set achieve better PDMS and EPDMS on NAVSIM navtest, with the gain concentrated in the product safety terms. Future work may explore better designs for the scorer training dataset and develop a better scoring module to capture the multi-modality of driving behavior.

{
    \small
    \bibliographystyle{ieeenat_fullname}
    \bibliography{main}

@String(PAMI = {IEEE Trans. Pattern Anal. Mach. Intell.})

@String(CVPR= {IEEE Conf. Comput. Vis. Pattern Recog.})

@String(ICCV= {Int. Conf. Comput. Vis.})

@String(ECCV= {Eur. Conf. Comput. Vis.})

@String(NIPS= {Adv. Neural Inform. Process. Syst.})

@String(ICLR = {Int. Conf. Learn. Represent.})

@String(AAAI = {AAAI})

@String(PAMI  = {IEEE TPAMI})

@String(CVPR  = {CVPR})

@String(ICCV  = {ICCV})

@String(ECCV  = {ECCV})

@String(NIPS  = {NeurIPS})

@String(ICLR  = {ICLR})

@inproceedings{dauner2024navsim,
  title     = {{NAVSIM}: Data-Driven Non-Reactive Autonomous Vehicle Simulation and Benchmarking},
  author    = {Dauner, Daniel and Hallgarten, Marcel and Li, Tianyu and Weng, Xinshuo and
               Huang, Zhiyu and Yang, Zetong and Li, Hongyang and Gilitschenski, Igor and
               Ivanovic, Boris and Pavone, Marco and Geiger, Andreas and Chitta, Kashyap},
  booktitle = NIPS,
  year      = {2024}
}

@inproceedings{dauner2023parting,
  title     = {Parting with Misconceptions about Learning-based Vehicle Motion Planning},
  author    = {Dauner, Daniel and Hallgarten, Marcel and Geiger, Andreas and Chitta, Kashyap},
  booktitle = {Conference on Robot Learning (CoRL)},
  year      = {2023}
}

@article{caesar2021nuplan,
  title   = {{nuPlan}: A Closed-Loop {ML}-Based Planning Benchmark for Autonomous Vehicles},
  author  = {Caesar, Holger and Kabzan, Juraj and Tan, Kok Seang and Fong, Whye Kit and
             Wolff, Eric and Lang, Alex and Fletcher, Luke and Beijbom, Oscar and Omari, Sammy},
  journal = {arXiv preprint arXiv:2106.11810},
  year    = {2021}
}

@article{chitta2023transfuser,
  title   = {{TransFuser}: Imitation with Transformer-Based Sensor Fusion for Autonomous Driving},
  author  = {Chitta, Kashyap and Prakash, Aditya and Jaeger, Bernhard and Yu, Zehao and
             Renz, Katrin and Geiger, Andreas},
  journal = PAMI,
  year    = {2023}
}

@inproceedings{hu2023uniad,
  title     = {Planning-oriented Autonomous Driving},
  author    = {Hu, Yihan and Yang, Jiazhi and Chen, Li and Li, Keyu and Sima, Chonghao and
               Zhu, Xizhou and Chai, Siqi and Du, Senyao and Lin, Tianwei and Wang, Wenhai and
               Lu, Lewei and Jia, Xiaosong and Liu, Qiang and Dai, Jifeng and Qiao, Yu and Li, Hongyang},
  booktitle = CVPR,
  year      = {2023}
}

@inproceedings{jiang2023vad,
  title     = {{VAD}: Vectorized Scene Representation for Efficient Autonomous Driving},
  author    = {Jiang, Bo and Chen, Shaoyu and Xu, Qing and Liao, Bencheng and Chen, Jiajie and
               Zhou, Helong and Zhang, Qian and Liu, Wenyu and Huang, Chang and Wang, Xinggang},
  booktitle = ICCV,
  year      = {2023}
}

@article{chen2024vadv2,
  title   = {{VADv2}: End-to-End Vectorized Autonomous Driving via Probabilistic Planning},
  author  = {Chen, Shaoyu and Jiang, Bo and Gao, Hao and Liao, Bencheng and Xu, Qing and
             Zhang, Qian and Huang, Chang and Liu, Wenyu and Wang, Xinggang},
  journal = {arXiv preprint arXiv:2402.13243},
  year    = {2024}
}

@inproceedings{liao2025diffusiondrive,
  title     = {{DiffusionDrive}: Truncated Diffusion Model for End-to-End Autonomous Driving},
  author    = {Liao, Bencheng and Chen, Shaoyu and Yin, Haoran and Jiang, Bo and Wang, Cheng and
               Yan, Sixu and Zhang, Xinbang and Li, Xiangyu and Zhang, Ying and Zhang, Qian and
               Wang, Xinggang},
  booktitle = CVPR,
  year      = {2025}
}

@article{li2024hydramdp,
  title     = {Hydra-mdp: End-to-end multimodal planning with multi-target hydra-distillation},
  author    = {Li, Zhenxin and Li, Kailin and Wang, Shihao and Lan, Shiyi and Yu, Zhiding and Ji, Yishen and Li, Zhiqi and Zhu, Ziyue and Kautz, Jan and Wu, Zuxuan and others},
  journal  = {arXiv preprint arXiv:2406.06978},
  year     = {2024}
}

@article{li2025gtrs,
  title   = {Generalized Trajectory Scoring for End-to-End Multimodal Planning},
  author  = {Li, Zhenxin and Yao, Wenhao and Wang, Zi and Sun, Xinglong and Chen, Joshua and
             Chang, Nadine and Shen, Maying and Wu, Zuxuan and Lan, Shiyi and Alvarez, Jose M.},
  journal = {arXiv preprint arXiv:2506.06664},
  year    = {2025}
}

@inproceedings{kirby2026drivor,
  title     = {Driving on Registers},
  author    = {Kirby, Ellington and Boulch, Alexandre and Xu, Yihong and Yin, Yuan and
               Puy, Gilles and Zablocki, {\'E}loi and Bursuc, Andrei and Gidaris, Spyros and
               Marlet, Renaud and Bartoccioni, Florent and Cao, Anh-Quan and Samet, Nermin and
               Vu, Tuan-Hung and Cord, Matthieu},
  booktitle = CVPR,
  year      = {2026}
}

@inproceedings{lin2017focal,
  title     = {Focal Loss for Dense Object Detection},
  author    = {Lin, Tsung-Yi and Goyal, Priya and Girshick, Ross and He, Kaiming and Doll{\'a}r, Piotr},
  booktitle = ICCV,
  year      = {2017}
}

@inproceedings{kumar2020cql,
  title     = {Conservative {Q}-Learning for Offline Reinforcement Learning},
  author    = {Kumar, Aviral and Zhou, Aurick and Tucker, George and Levine, Sergey},
  booktitle = NIPS,
  year      = {2020}
}

@inproceedings{kostrikov2022iql,
  title     = {Offline Reinforcement Learning with Implicit {Q}-Learning},
  author    = {Kostrikov, Ilya and Nair, Ashvin and Levine, Sergey},
  booktitle = ICLR,
  year      = {2022}
}

@inproceedings{tian2026simscale,
  title     = {SimScale: Learning to Drive via Real-World Simulation at Scale},
  author    = {Tian, Haochen and Li, Tianyu and Liu, Haochen and Yang, Jiazhi
               and Qiu, Yihang and Li, Guang and Wang, Junli and Gao, Yinfeng
               and Zhang, Zhang and Wang, Liang and Ye, Hangjun and Tan, Tieniu
               and Chen, Long and Li, Hongyang},
  booktitle = {CVPR},
  year      = {2026}
}

@inproceedings{goalflow,
  title     = {{GoalFlow}: Goal-Driven Flow Matching for Multimodal Trajectories Generation in End-to-End Autonomous Driving},
  author    = {Xing, Zebin and Zhang, Xingyu and Hu, Yang and Jiang, Bo and He, Tong and
               Zhang, Qian and Long, Xiaoxiao and Yin, Wei},
  booktitle = CVPR,
  year      = {2025}
}

@inproceedings{wote,
  title     = {End-to-End Driving with Online Trajectory Evaluation via {BEV} World Model},
  author    = {Li, Yingyan and Wang, Yuqi and Liu, Yang and He, Jiawei and Fan, Lue and
               Zhang, Zhaoxiang},
  booktitle = ICCV,
  year      = {2025}
}

@inproceedings{world4drive,
  title     = {{World4Drive}: End-to-End Autonomous Driving via Intention-aware Physical Latent World Model},
  author    = {Zheng, Yupeng and Yang, Pengxuan and Xing, Zebin and Zhang, Qichao and
               Zheng, Yuhang and Gao, Yinfeng and Li, Pengfei and Zhang, Teng and
               Xia, Zhongpu and Jia, Peng and Lang, XianPeng and Zhao, Dongbin},
  booktitle = ICCV,
  year      = {2025}
}

@inproceedings{epona,
  title     = {{Epona}: Autoregressive Diffusion World Model for Autonomous Driving},
  author    = {Zhang, Kaiwen and Tang, Zhenyu and Hu, Xiaotao and Pan, Xingang and
               Guo, Xiaoyang and Liu, Yuan and Huang, Jingwei and Yuan, Li and
               Zhang, Qian and Long, Xiao-Xiao and Cao, Xun and Yin, Wei},
  booktitle = ICCV,
  year      = {2025}
}

@article{hydramdppp,
  title   = {{Hydra-MDP++}: Advancing End-to-End Driving via Expert-Guided Hydra-Distillation},
  author  = {Li, Kailin and Li, Zhenxin and Lan, Shiyi and Xie, Yuan and Zhang, Zhizhong and
             Liu, Jiayi and Wu, Zuxuan and Yu, Zhiding and Alvarez, Jose M.},
  journal = {arXiv preprint arXiv:2503.12820},
  year    = {2025}
}

@inproceedings{drivesuprim,
  title     = {{DriveSuprim}: Towards Precise Trajectory Selection for End-to-End Planning},
  author    = {Yao, Wenhao and Li, Zhenxin and Lan, Shiyi and Wang, Zi and Sun, Xinglong and
               Alvarez, Jose M. and Wu, Zuxuan},
  booktitle = AAAI,
  year      = {2026}
}

@inproceedings{meanfuser,
  title    = {Meanfuser: Fast one-step multi-modal trajectory generation and adaptive reconstruction via meanflow for end-to-end autonomous driving},
  author   = {Wang, Junli and Zheng, Yinan and Liu, Xueyi and Xing, Zebin and Li, Pengfei and Ma, Kun and Ye, Hangjun and Chen, Guang and Li, Guang and Chen, Long and others},
  booktitle = CVPR,
  year      = {2026}
}

@inproceedings{ross2011dagger,
  title     = {A Reduction of Imitation Learning and Structured Prediction to
               No-Regret Online Learning},
  author    = {Ross, St{\'e}phane and Gordon, Geoffrey J. and Bagnell, J. Andrew},
  booktitle = {Proc. Int. Conf. Artificial Intelligence and Statistics (AISTATS)},
  year      = {2011}
}

@inproceedings{bansal2019chauffeurnet,
  title     = {{ChauffeurNet}: Learning to Drive by Imitating the Best and
               Synthesizing the Worst},
  author    = {Bansal, Mayank and Krizhevsky, Alex and Ogale, Abhijit},
  booktitle = {Robotics: Science and Systems (RSS)},
  year      = {2019}
}

@inproceedings{he2016resnet,
  title     = {Deep residual learning for image recognition},
  author    = {He, Kaiming and Zhang, Xiangyu and Ren, Shaoqing and Sun, Jian},
  booktitle = CVPR,
  year      = {2016}
}

@inproceedings{cao2025navsim_v2,
  title     = {Pseudo-Simulation for Autonomous Driving},
  author    = {Cao, Wei and Hallgarten, Marcel and Li, Tianyu and Dauner, Daniel
               and Gu, Xunjiang and Wang, Caojun and Miron, Yakov and Aiello, Marco
               and Li, Hongyang and Gilitschenski, Igor and Ivanovic, Boris
               and Pavone, Marco and Geiger, Andreas and Chitta, Kashyap},
  booktitle = {Conference on Robot Learning (CoRL)},
  year      = {2025}
}

@article{li2025ztrs,
  title   = {Ztrs: Zero-imitation end-to-end autonomous driving with trajectory scoring},
  author  = {Li, Zhenxin and Yao, Wenhao and Wang, Zi and Sun, Xinglong and Chen, Jingde and Chang, Nadine and Shen, Maying and Song, Jingyu and Wu, Zuxuan and Lan, Shiyi and others},
  journal = {arXiv preprint arXiv:2510.24108},
  year    = {2025}
}

@article{park2023hiql,
  title   = {Hiql: Offline goal-conditioned rl with latent states as actions},
  author  = {Park, Seohong and Ghosh, Dibya and Eysenbach, Benjamin and Levine, Sergey},
  journal = NIPS,
  year   = {2023}
}

@article{zhou2026hugsim,
  title   = {Hugsim: A real-time, photo-realistic and closed-loop simulator for autonomous driving},
  author  = {Zhou, Hongyu and Lin, Longzhong and Wang, Jiabao and Lu, Yichong and Bai, Dongfeng and Liu, Bingbing and Wang, Yue and Geiger, Andreas and Liao, Yiyi},
  journal = PAMI,
  year    = {2026}
}

@inproceedings{nuplanbench,
  title     = {Towards Learning-Based Planning: The nuPlan Benchmark for
               Real-World Autonomous Driving},
  author    = {Karnchanachari, Napat and Geromichalos, Dimitris and Tan, Kok Seang
               and Li, Nanxiang and Eriksen, Christopher and Yaghoubi, Shakiba
               and Mehdipour, Noushin and Bernasconi, Gianmarco and Fong, Whye Kit
               and Guo, Yiluan and Caesar, Holger},
  booktitle = {IEEE International Conference on Robotics and Automation (ICRA)},
  year      = {2024}
}

@inproceedings{egostatus,
  title     = {Is Ego Status All You Need for Open-Loop End-to-End
               Autonomous Driving?},
  author    = {Li, Zhiqi and Yu, Zhiding and Lan, Shiyi and Li, Jiahan
               and Kautz, Jan and Lu, Tong and Alvarez, Jose M.},
  booktitle = CVPR,
  year      = {2024}
}

@inproceedings{hiddenbiases,
  title     = {Hidden Biases of End-to-End Driving Models},
  author    = {Jaeger, Bernhard and Chitta, Kashyap and Geiger, Andreas},
  booktitle = ICCV,
  year      = {2023}
}

@inproceedings{behaviorcloning,
  title     = {Exploring the Limitations of Behavior Cloning for
               Autonomous Driving},
  author    = {Codevilla, Felipe and Santana, Eder and L\'opez, Antonio M.
               and Gaidon, Adrien},
  booktitle = ICCV,
  year      = {2019}
}

@article{galashov2022data,
  title   = {Data augmentation for efficient learning from parametric experts},
  author  = {Galashov, Alexandre and Merel, Josh S and Heess, Nicolas},
  journal = NIPS,
  year    = {2022}
}

@inproceedings{dataaggregation,
  title     = {Exploring Data Aggregation in Policy Learning for
               Vision-Based Urban Autonomous Driving},
  author    = {Prakash, Aditya and Behl, Aseem and Ohn-Bar, Eshed
               and Chitta, Kashyap and Geiger, Andreas},
  booktitle = CVPR,
  year      = {2020}
}

@inproceedings{cheating,
  title     = {Learning by Cheating},
  author    = {Chen, Dian and Zhou, Brady and Koltun, Vladlen
               and Kr\"ahenb\"uhl, Philipp},
  booktitle = {Conference on Robot Learning (CoRL)},
  year      = {2020}
}

@inproceedings{king,
  title     = {KING: Generating Safety-Critical Driving Scenarios for
               Robust Imitation via Kinematics Gradients},
  author    = {Hanselmann, Niklas and Renz, Katrin and Chitta, Kashyap
               and Bhattacharyya, Apratim and Geiger, Andreas},
  booktitle = ECCV,
  year      = {2022}
}

@inproceedings{counterfactual,
  title     = {Counterfactual Data Augmentation Using Locally Factored
               Dynamics},
  author    = {Pitis, Silviu and Creager, Elliot and Garg, Animesh},
  booktitle = NIPS,
  year      = {2020}
}

@inproceedings{unisim,
  title     = {UniSim: A Neural Closed-Loop Sensor Simulator},
  author    = {Yang, Ze and Chen, Yun and Wang, Jingkang
               and Manivasagam, Sivabalan and Ma, Wei-Chiu
               and Yang, Anqi Joyce and Urtasun, Raquel},
  booktitle = CVPR,
  year      = {2023}
}

@inproceedings{neurad,
  title     = {NeuRAD: Neural Rendering for Autonomous Driving},
  author    = {Tonderski, Adam and Lindstr\"om, Carl and Hess, Georg
               and Ljungbergh, William and Svensson, Lennart
               and Petersson, Christoffer},
  booktitle = CVPR,
  year      = {2024}
}

@inproceedings{sparsedrive,
  title     = {SparseDrive: End-to-End Autonomous Driving via Sparse
               Scene Representation},
  author    = {Sun, Wenchao and Lin, Xuewu and Shi, Yining and Zhang, Chuang
               and Wu, Haoran and Zheng, Sifa},
  booktitle = {IEEE International Conference on Robotics and Automation (ICRA)},
  year      = {2025}
}
}

\end{document}